\documentclass[runningheads]{llncs}
\usepackage[T1]{fontenc}
\usepackage{multirow}
\usepackage{makecell}
\usepackage{booktabs}
\usepackage{graphicx}
\usepackage{amsmath}
\usepackage{amssymb}
\usepackage{subcaption}
\usepackage{wrapfig}
\usepackage{marvosym}
\usepackage[colorlinks=true,citecolor=blue,urlcolor=blue,linkcolor=blue]{hyperref}

\usepackage{amsmath}
\begin{document}
\title{Hierarchical Possession-Aware Graph Pointer Network for Pass Receiver Selection}
\titlerunning{HPGPN for Pass Receiver Selection}
\author{Jingyi Wang\inst{1,2}\orcidID{0009-0007-6112-8373} \and Da Li\inst{1,2} \and
Kaixin Wang\inst{1,2}\textsuperscript{(*)}\orcidID{0000-0002-6650-2850} \and Zhangqin Huang\inst{1,2}\textsuperscript{(*)}
}

\authorrunning{J. Wang et al.}
\institute{Beijing Engineering Research Center for IoT Software and Systems, Beijing University of Technology, Beijing, China
 \and Beijing Key Laboratory of Interdisciplinary Intelligent Technologies in Sports Medicine and Engineering, Beijing University of Technology, Beijing, China\\
 \email{jingyiwang@emails.bjut.edu.cn}, \email{\{lida1204,kaixin.wang,zhuang\}@bjut.edu.cn}}
%\institute{Princeton University, Princeton NJ 08544, USA \and
%Springer Heidelberg, Tiergartenstr. 17, 69121 Heidelberg, Germany
\maketitle             % typeset the header of the contribution
\begingroup
\renewcommand{\thefootnote}{*}
\footnotetext{K. Wang and Z. Huang---Equal contributors and co-corresponding authors.}
\endgroup

\begin{abstract}

Pass receiver selection is a fundamental task in football analytics, aiming to predict the intended receiver under a given game state. This task is challenging with event-centered freeze-frame observations, a broadcast-like setting that provides only partial and variable player visibility without complete trajectories or stable player identities. The model must therefore reason over anonymous visible candidates, opponent pressure, and recent context under partial observation. To address this setting, we propose a Hierarchical Possession-aware Graph Pointer Network (HPGPN), which formulates pass receiver selection as variable-size candidate prediction over visible teammates. HPGPN jointly models current player interactions, local event context, and possession-level temporal dynamics. It represents the current pass situation with a graph, incorporates fixed event context, and uses dynamic possession history to capture how the attacking sequence evolves. Candidate representations are refined hierarchically by integrating spatial, contextual, and historical evidence, and a glimpse pointer head scores the receiver candidates. Experiments on public football event and freeze-frame data show that HPGPN improves pass receiver selection performance. Ablation studies demonstrate the effectiveness of graph-based interaction modeling, fixed event context, and dual-branch dynamic possession-history modeling.

\end{abstract}
\section{Introduction}

Passing is one of the most frequent and consequential decisions in football, as it changes possession structure and shapes subsequent attacking opportunities~\cite{fujii2025machine}. 
A key problem in modeling passing behavior is pass receiver selection: given the current game state, which teammate will the ball carrier choose as the intended receiver? This problem requires reasoning over multiple plausible teammates, opponent pressure, spatial availability, and the tactical context.

Existing studies have explored pass receiver prediction using tracking data or video observations, showing that receiver choice depends heavily on spatial and temporal player configurations~\cite{honda2022pass,kaneko2024augmenting,rahimian2026temporal,vercruyssen2016qualitative}.
However, these settings often require dense trajectories, visual preprocessing, or richer annotations. In contrast, un-xPass~\cite{robberechts2023xpass} addresses this issue using event-centered freeze-frame observations from StatsBomb 360 Data\footnote{StatsBomb Open Data is available at \url{https://github.com/statsbomb/open-data}.}. 
This broadcast-like setting provides partial player-location snapshots around specific events and can be scaled to a wider range of competitions. 
In this setting, the visible player set varies across events, non-event players are not consistently identifiable across snapshots, and receiver selection is defined over a variable-size set of anonymous visible teammates.

A natural limitation of modeling each pass only from the current freeze frame in the un-xPass setting is that the receiver choice is not determined solely by the instantaneous spatial layout. The intended receiver is usually affected by preceding actions, short-term passing patterns, attacking direction, and possession evolution. Moreover, candidates should not be treated as independent class labels, since each candidate's selection likelihood depends on interactions with the passer, nearby opponents, event context, and possession history.

To address these challenges, we propose a Hierarchical Possession-aware Graph Pointer Network (HPGPN) for pass receiver selection from event-centered freeze-frame observations. 
HPGPN represents the current pass as a player interaction graph, incorporates fixed event context, and encodes preceding possession events through a dual-branch history encoder. HPGPN follows a hierarchical decision architecture. It progressively enriches receiver candidates with graph-based interactions, local event context, and historical possession evidence, while also constructing a pass-level decision query from graph, event, and history representations. A glimpse pointer head then refines this query by attending to the available candidate set and produces normalized scores over the current receiver candidates. This design allows the model to reason over variable-size sets of visible receiver candidates without relying on a fixed global player vocabulary while accounting for how the attacking sequence has developed before the pass. 

Our contributions are summarized as follows:
\begin{itemize}
    \item[$\bullet$] We introduce a possession-aware formulation for pass receiver selection from event-centered freeze-frame observations, emphasizing the role of both current spatial configuration and preceding possession development.
    
    \item[$\bullet$] We propose HPGPN, a hierarchical possession-aware graph pointer network that integrates player-interaction modeling, event-level context, and possession-history reasoning.
    
    \item[$\bullet$] We design a dual-branch possession history encoder, hierarchical candidate refinement, and a glimpse pointer head to capture complementary temporal evidence and directly score variable-size receiver candidate sets.
    
    \item[$\bullet$] We conduct extensive experiments and ablation studies on public football datasets to validate the effectiveness, generalization ability, and component contributions of the proposed model.
\end{itemize}

\section{Related Work}
\subsection{Pass Prediction and Pass Receiver Selection}

Pass prediction has been studied as a core problem in football analytics, with early work influenced by the MLSA 2018 pass prediction challenge~\cite{dauxais2018predicting,fournier2018football,hubavcek2018deep,li2018predicting}. In this challenge, the goal is to predict the receiver of a pass given the pass sender and the locations of all players on the pitch at the time of the pass. Previous approaches~\cite{dauxais2018predicting,fournier2018football} relied on hand-crafted spatial features, such as distances and angles among the passer, candidate receivers, and nearby opponents. Li and Zhang~\cite{li2018predicting} formulated receiver prediction as a learning-to-rank problem, which aligns naturally with selecting one receiver from a candidate set. Hubáček et al.~\cite{hubavcek2018deep} further explored neural representations of player configurations. These studies established the importance of local geometric relations.

Subsequent work has incorporated richer data sources, including tracking data, video frames, and multimodal inputs. Tracking-based approaches enable fine-grained modeling of continuous player movement and defensive organization \cite{vercruyssen2016qualitative,rahimian2026temporal}. Video-based methods predict pass recipients from broadcast or wide-angle visual inputs, often combined with annotations, player trajectories, or top-view coordinates \cite{sanyal2021will,honda2022pass,paneru2024enhancing,kaneko2024augmenting}. Although these methods demonstrate the value of spatial, temporal, and contextual information, they often depend on full tracking data, visual preprocessing, stable player identities, or additional annotations, which can limit scalability across competitions.

Among prior methods, SoccerMap~\cite{fernandez2020soccermap}  and un-xPass~\cite{robberechts2023xpass} are particularly relevant to our experimental setting. 
SoccerMap introduced a visually interpretable deep learning framework based on pitch-level spatial maps, showing the effectiveness of structured spatial representations for modeling passing decisions. 
More recently, un-xPass introduced a pass creativity metric using StatsBomb Open Data. As one component, it estimates pass receiver selection and evaluates receiver-based baselines, including an XGBoost ranking model built on handcrafted spatial features for visible pass options. We follow the un-xPass setting, where receiver selection is performed over a variable-size set of anonymous visible teammates, but shift the focus to pass receiver selection as a standalone modeling problem and develop a structured neural approach beyond handcrafted feature-based ranking.

\subsection{Graph Pointer Networks}

Pointer Networks~\cite{vinyals2015pointer} were introduced to handle output spaces whose size depends on the input by using attention to select elements from the input sequence. 
They have been widely used in structured prediction and combinatorial optimization problems, where the output can be represented as an ordering or selection over input elements. Graph neural networks provide a natural way to encode relational structure among input elements. 
Representative architectures include graph convolutional network (GCN)~\cite{kipf2017semi}, GraphSAGE~\cite{hamilton2017inductive}, and graph attention network (GAT)~\cite{velivckovic2018graph}.

Combining these two ideas, Graph Pointer Networks~\cite{ma2019combinatorial} extend Pointer Networks by introducing a graph embedding layer before the pointer decoder, so that relationships between input nodes can be captured before node selection. 
Our work follows the same paradigm of combining graph-based relational encoding with pointer-style candidate selection, but applies it to a different decision-making problem.

\section{Problem Formulation}
\label{sec:problem_formulation}

Given a pass event $p$, pass receiver selection aims to predict the intended receiver from the visible teammate candidates. 
Since the visible teammate set varies across events, the task is formulated as candidate selection over a pass specific and variable size set, rather than classification over a fixed global player vocabulary. For each pass $p$, we observe an event-centered freeze frame containing the passer, visible teammates, visible opponents, and their spatial information. 
The candidate receiver set is defined as
\[
\mathcal{C}_p = \{c_1, c_2, \ldots, c_{N_p}\},
\]
where $N_p$ denotes the number of visible teammate candidates for pass $p$, excluding the passer. 
The ground-truth receiver is represented by
\[
y_p \in \{1,\ldots,N_p\},
\]
where $y_p=n$ indicates that candidate $c_n$ is the observed receiver. 
The objective is to estimate a categorical distribution over the current candidate set:
\[
P(y_p=n \mid p,\mathcal{C}_p), \quad n \in \{1,\ldots,N_p\},
\]
where $p$ includes all available pre-pass inputs, such as the current pass graph, fixed event context, and dynamic possession history. All inputs are restricted to information available before or at the pass moment. We exclude the true receiver, pass outcome, post-pass movement, and subsequent events. Historical outcomes are used only for events preceding the current pass. 

As depicted in Fig.~\ref{fig:overview}, each current pass event $a_t$ is associated with a StatsBomb 360 freeze frame and an event sequence in the same possession:
\[
(a_{t-L}, a_{t-L+1}, \ldots, a_{t-1}, a_t),
\]
where $a_{t-i}$ denotes the $i$-th preceding event before the current pass and $L$ is bounded by the maximum history window length.

The predicted receiver is obtained as
\[
\hat{y}_p =
\arg\max_{n \in \{1,\ldots,N_p\}}
P(y_p=n \mid p,\mathcal{C}_p).
\]

\section{Methodology}
\label{sec:methodology}

\begin{figure}[t]
\centering
\vspace{-0.5cm}
\includegraphics[width=\textwidth]{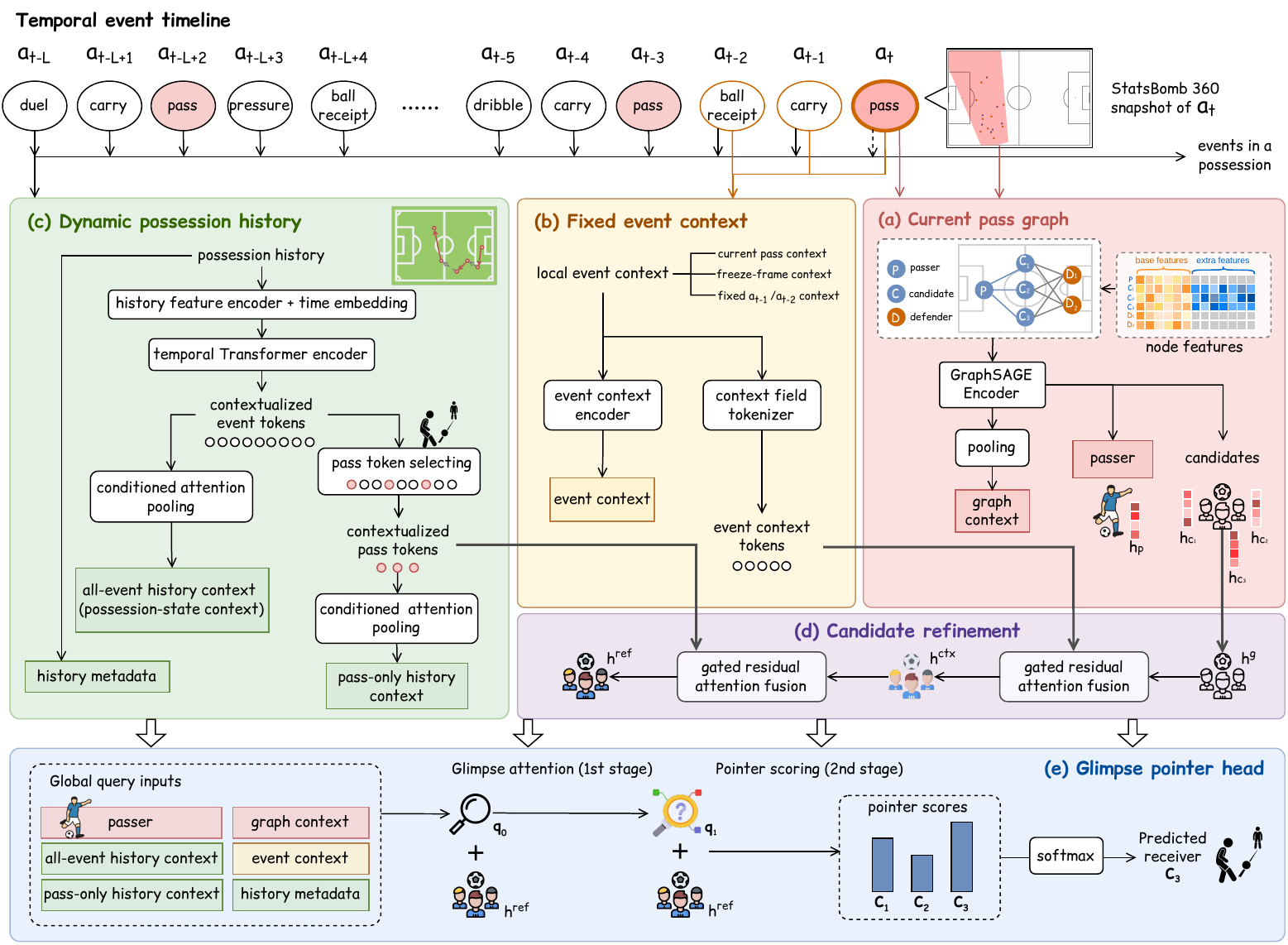} % Reduce the figure size so that it is slightly narrower than the column.
\caption{The overview of the proposed HPGPN framework. (a) Current pass graph for player interactions; (b) Fixed event context for local event-level cues; (c) Dual-branch dynamic possession history for all-event and pass-only contexts; (d) Hierarchical candidate refinement for representation integration; (e) Glimpse pointer head for variable-size receiver selection.}
\label{fig:overview}
% \Description{}
\vspace{-0.5cm}
\end{figure}

The overall framework of HPGPN is illustrated in Fig.~\ref{fig:overview}. HPGPN models pass receiver selection through a hierarchical possession-aware graph pointer architecture, with a candidate-centric design that progressively refines receiver representations and the pass-level decision query. Specifically, (1) the current pass is structured as a pass-centric graph that explicitly models passing options and defensive interactions among the passer, candidate receivers, and defenders; (2) the fixed event context provides current event-level and short-range contextual cues; (3) preceding possession events are encoded into a shared contextualized history representation, from which a dual-branch design derives an all-event context for possession evolution and a pass-only context for historical passing patterns. Then, the representations are integrated through hierarchical candidate refinement, where each candidate is initialized from the current pass graph, conditioned on fixed event context, and further enriched with dynamic possession history. A pass-level decision query is also constructed to capture the overall decision context. Finally, the glimpse pointer head first summarizes the current candidate set to refine the decision query and then produces pointer scores over the variable-size candidate set to predict the most likely pass receiver.

\subsection{Current Pass Graph Encoder}
\label{subsec:current_pass_graph_encoder}

The current pass graph encoder captures the instantaneous relational structure of the pass situation. 
For each pass event $p$, we construct a graph
\[
\mathcal{G}=(\mathcal{V},\mathcal{E}),
\]
where $\mathcal{V}$ contains the passer, visible candidate receivers, and visible opponents. 
Opponents are included as defender nodes to provide defensive context. 
We add directed edges from the passer to each candidate receiver and bidirectional edges between candidate receivers and defenders.

Each node is initialized with spatial features and node-type indicators. 
Candidate receiver nodes additionally include pass-option features such as pass distance, pass angle, and defensive proximity~\cite{robberechts2023xpass}. 
Given the initial node feature $\mathbf{x}_v$, we project it into the hidden space and apply a GraphSAGE encoder:
\[
\mathbf{h}_v^{g} =
\operatorname{GraphSAGE}
\left(
\operatorname{MLP}_{\mathrm{node}}(\mathbf{x}_v), \mathcal{E}
\right),
\]

From the encoded graph, we extract the passer representation from the passer node $v_0$ and stack the graph-encoded candidate receiver representations as
\[
\mathbf{h}_{\mathrm{passer}} = \mathbf{h}_{v_0}^{g},
\qquad
\mathbf{H}_{c}^{g}
=
[\mathbf{h}_{c_1}^{g};\mathbf{h}_{c_2}^{g};\ldots;\mathbf{h}_{c_{N_p}}^{g}]
\in \mathbb{R}^{N_p \times d},
\]
A graph-level context vector is computed by mean pooling over all graph nodes:
\[
\mathbf{h}_{\mathrm{graph}}
=
\frac{1}{|\mathcal{V}|}
\sum_{v \in \mathcal{V}}
\mathbf{h}_v^{g}.
\]
The resulting representations $\mathbf{h}_{\mathrm{passer}}$, $\mathbf{H}_{c}^{g}$, and $\mathbf{h}_{\mathrm{graph}}$ summarize the current spatial decision context and are used by the subsequent context encoding and candidate refinement modules.

\subsection{Fixed Event Context Encoder}
\label{subsec:fixed_event_context_encoder}
While the current pass graph models player-level relations, it does not explicitly encode compact event-level cues shared by all receiver candidates. 
We therefore introduce a fixed event context encoder for event-level information. 
For each pass event $p$ corresponding to the current event $a_t$, the fixed event context vector $\mathbf{x}_{\mathrm{ctx}}$ includes current pass attributes, freeze-frame summary features, and short-range gamestate information from the two immediately preceding actions $a_{t-2}$ and $a_{t-1}$. These preceding actions are not restricted to passes.

We encode the full context vector into a global event-context representation:
\[
\mathbf{h}_{\mathrm{ctx}}
=
\operatorname{MLP}_{\mathrm{ctx}}
(
\mathbf{x}_{\mathrm{ctx}}
),
\quad
\mathbf{h}_{\mathrm{ctx}} \in \mathbb{R}^{d}.
\]

To support candidate-specific context retrieval, we also convert each scalar context feature into a token. For the $j$-th context feature $x_{\mathrm{ctx},j}$, we compute
\[
\mathbf{t}^{\mathrm{ctx}}_j
=
\operatorname{MLP}_{\mathrm{val}}
(
x_{\mathrm{ctx},j}
)
+
\mathbf{e}^{\mathrm{ctx}}_j,
\]
where $\operatorname{MLP}_{\mathrm{val}}$ maps the scalar value into the hidden space, and $\mathbf{e}^{\mathrm{ctx}}_j \in \mathbb{R}^{d}$ is a learnable field embedding. The fixed-context token sequence is
\[
\mathbf{T}_{\mathrm{ctx}}
=
[
\mathbf{t}^{\mathrm{ctx}}_1;
\ldots;
\mathbf{t}^{\mathrm{ctx}}_{D_{\mathrm{ctx}}}
]
\in \mathbb{R}^{D_{\mathrm{ctx}} \times d}.
\]

\subsection{Dual-Branch Dynamic Possession History Encoder}
\label{subsec:dual_branch_dynamic_history_encoder}

Receiver selection depends on both the current pass and the preceding possession development. We therefore introduce a dual-branch dynamic possession history encoder, which contextualizes preceding possession events and derives two complementary history contexts: an all-event context and a pass-only context.

For each current pass event $a_t$, we retain $L$ preceding events from the same possession, where $L$ is bounded by the maximum history window length $K$:
\[
\mathcal{S}_p = (s_1, s_2, \ldots, s_L),
\quad 0 \leq L \leq K,
\]
where
$s_j = a_{t-j}$. Thus, $\mathcal{S}_p$ is ordered from recent to early events and excludes $a_t$. 
Each event $s_j$ is represented by an event feature vector $\mathbf{x}_{j}^{h}$ and a temporal feature vector $\boldsymbol{\tau}_{j}$, where $\mathbf{x}_{j}^{h}$ describes the action attributes, spatial and pressure cues, available action outcomes, and $\boldsymbol{\tau}_{j}$ encodes the relative time difference to $a_t$, and recency position.

\subsubsection{Contextualized Possession-History Tokens}
\label{subsubsec:contextualized_history_tokens}

Each historical event $s_j \in \mathcal{S}_p$ is encoded as a time-aware token by combining its event features and temporal features:
\[
\mathbf{z}_{j}^{(0)}
=
f_h(\mathbf{x}_{j}^{h})
+
f_{\tau}(\boldsymbol{\tau}_{j}),
\quad j=1,\ldots,L,
\]
where $f_h$ and $f_{\tau}$ are learnable encoders. 
Stacking the tokens yields
\[
\mathbf{Z}^{(0)}
=
[
\mathbf{z}_{1}^{(0)};
\ldots;
\mathbf{z}_{L}^{(0)}
]
\in \mathbb{R}^{L \times d}.
\]
We contextualize the sequence using a temporal Transformer encoder:
\[
\mathbf{Z}
=
\operatorname{TransformerEncoder}
(
\mathbf{Z}^{(0)}
),
\quad
\mathbf{Z}
=
[
\mathbf{z}_{1};
\ldots;
\mathbf{z}_{L}
]
\in \mathbb{R}^{L \times d}.
\]
The shared contextualized sequence $\mathbf{Z}$ is used to derive both all-event and pass-only history contexts.

\subsubsection{All-Event History Context}
\label{subsubsec:all_event_history_context}

The all-event branch summarizes the overall possession development before the current pass by attending to all contextualized history tokens. 
We first construct a pass-conditioned query:
\[
\mathbf{q}_{\mathrm{all}}
=
\operatorname{MLP}_{\mathrm{all}}
\left(
[
\mathbf{h}_{\mathrm{passer}};
\mathbf{h}_{\mathrm{graph}};
\mathbf{h}_{\mathrm{ctx}};
\mathbf{m}_{\mathrm{all}}
]
\right),
\]
where $\mathbf{m}_{\mathrm{all}}$ encodes history availability and normalized history length. 
Attention scores over the contextualized tokens are computed as
\[
e_{j}^{\mathrm{all}}
=
\mathbf{v}_{\mathrm{all}}^{\top}
\tanh
\left(
\mathbf{W}_{\mathrm{all}}\mathbf{z}_{j}
+
\mathbf{U}_{\mathrm{all}}\mathbf{q}_{\mathrm{all}}
\right),
\quad j=1,\ldots,L.
\]
The weights are normalized over the retained history:
\[
\alpha_{j}^{\mathrm{all}}
=
\frac{\exp(e_{j}^{\mathrm{all}})}
{\sum_{r=1}^{L}\exp(e_{r}^{\mathrm{all}})},
\]
and the all-event history context is
\[
\mathbf{h}_{\mathrm{all}}
=
\sum_{j=1}^{L}
\alpha_{j}^{\mathrm{all}}\mathbf{z}_{j},
\quad
\mathbf{h}_{\mathrm{all}} \in \mathbb{R}^{d}.
\]
When $L=0$, $\mathbf{h}_{\mathrm{all}}$ is set to zero and the absence of history is indicated by $\mathbf{m}_{\mathrm{all}}$.

\subsubsection{Pass-Only History Context}
\label{subsubsec:pass_only_history_context}

The pass-only branch summarizes historical pass actions, which may indicate recent passing patterns and attacking direction. 
We define the pass-token index set as
$
\mathcal{P}_p
=
\{j \in \{1,\ldots,L\} \mid s_j \text{ is a pass action}\}
$.
The corresponding pass-token sequence is selected from the shared contextualized history representation as $\mathbf{Z}_{\mathrm{pass}}=\{\mathbf{z}_j \mid j \in \mathcal{P}_p\}$, so each selected token retains both pass-specific information and surrounding possession context.

We construct a pass-history query:
\[
\mathbf{q}_{\mathrm{pass}}
=
\operatorname{MLP}_{\mathrm{pass}}
\left(
[
\mathbf{h}_{\mathrm{passer}};
\mathbf{h}_{\mathrm{graph}};
\mathbf{h}_{\mathrm{ctx}};
\mathbf{m}_{\mathrm{dual}}
]
\right),
\]
where $\mathbf{m}_{\mathrm{dual}}$ encodes the availability and normalized lengths of both the full history and pass-only history. 
For each selected pass token $j \in \mathcal{P}_p$, we compute
\[
e_j^{\mathrm{pass}}
=
\mathbf{v}_{\mathrm{pass}}^{\top}
\tanh
\left(
\mathbf{W}_{\mathrm{pass}}\mathbf{z}_j
+
\mathbf{U}_{\mathrm{pass}}\mathbf{q}_{\mathrm{pass}}
\right),
\]
\[
\alpha_j^{\mathrm{pass}}
=
\frac{\exp(e_j^{\mathrm{pass}})}
{\sum_{r \in \mathcal{P}_p}\exp(e_r^{\mathrm{pass}})},
\quad
\mathbf{h}_{\mathrm{pass}}
=
\sum_{j \in \mathcal{P}_p}
\alpha_j^{\mathrm{pass}}\mathbf{z}_j.
\]
If $\mathcal{P}_p=\emptyset$, $\mathbf{h}_{\mathrm{pass}}$ is set to zero, with the absence indicated by $\mathbf{m}_{\mathrm{dual}}$.

\subsection{Hierarchical Candidate Refinement}
\label{subsec:hierarchical_candidate_refinement}

HPGPN refines each candidate representation in a hierarchical manner. 
Starting from the graph-based candidate embedding $\mathbf{h}_{c_n}^{g}$, the refinement module sequentially injects fixed event context and candidate-specific pass-history evidence, yielding the final representation $\mathbf{h}_{c_n}^{\mathrm{ref}}$.

Given the fixed-context token sequence $\mathbf{T}_{\mathrm{ctx}}$, candidate $c_n$ first retrieves local event context by attention:
\[
\mathbf{a}_{n}^{\mathrm{ctx}}
=
\operatorname{CrossAttention}
\left(
\mathbf{h}_{c_n}^{g},
\mathbf{T}_{\mathrm{ctx}},
\mathbf{T}_{\mathrm{ctx}}
\right).
\]
The retrieved context is integrated through gated residual fusion:
\[
\mathbf{g}_{n}^{\mathrm{ctx}}
=
\sigma
\left(
\operatorname{MLP}_{g}^{\mathrm{ctx}}
\left(
[
\mathbf{h}_{c_n}^{g};
\mathbf{a}_{n}^{\mathrm{ctx}}
]
\right)
\right),
\quad
\mathbf{h}_{c_n}^{\mathrm{ctx}}
=
\operatorname{LayerNorm}
\left(
\mathbf{h}_{c_n}^{g}
+
\mathbf{g}_{n}^{\mathrm{ctx}}
\odot
\mathbf{a}_{n}^{\mathrm{ctx}}
\right).
\]

We then refine the candidate using pass-history tokens selected from the contextualized possession history:
\[
\mathbf{a}_{n}^{\mathrm{pass}}
=
\operatorname{CrossAttention}
\left(
\mathbf{h}_{c_n}^{\mathrm{ctx}},
\mathbf{Z}_{\mathrm{pass}},
\mathbf{Z}_{\mathrm{pass}}
\right).
\]
The pass-history message is fused with another gated residual update:
\[
\mathbf{g}_{n}^{\mathrm{pass}}
=
\sigma
\left(
\operatorname{MLP}_{g}^{\mathrm{pass}}
\left(
[
\mathbf{h}_{c_n}^{\mathrm{ctx}};
\mathbf{a}_{n}^{\mathrm{pass}}
]
\right)
\right),
\]
\[
\mathbf{h}_{c_n}^{\mathrm{ref}}
=
\operatorname{LayerNorm}
\left(
\mathbf{h}_{c_n}^{\mathrm{ctx}}
+
\tanh(\lambda_{\mathrm{pass}})
\,
\mathbf{g}_{n}^{\mathrm{pass}}
\odot
\mathbf{a}_{n}^{\mathrm{pass}}
\right).
\]
Here, $\lambda_{\mathrm{pass}}$ is a learnable scalar initialized to zero. 
The final representation $\mathbf{h}_{c_n}^{\mathrm{ref}}$ integrates graph-based spatial relations, fixed event context, and candidate-specific pass-history evidence.

\subsection{Glimpse Pointer Receiver Selection Head}
\label{subsec:glimpse_pointer_head}

After hierarchical refinement, each candidate $c_n$ is represented by $\mathbf{h}_{c_n}^{\mathrm{ref}}$. 
Since the number of candidates varies across pass events, we use a glimpse pointer head to score the current candidate set.

We first construct an initial pointer query from the pass-level context:
\[
\mathbf{q}^{(0)}
=
\operatorname{MLP}_{q}
\left(
[
\mathbf{h}_{\mathrm{passer}};
\mathbf{h}_{\mathrm{graph}};
\mathbf{h}_{\mathrm{ctx}};
\mathbf{h}_{\mathrm{all}};
\mathbf{h}_{\mathrm{pass}};
\mathbf{m}_{\mathrm{dual}}
]
\right),
\]

The head then performs a glimpse attention step over the refined candidate embeddings:
\[
e_n^{\mathrm{glp}}
=
\mathbf{v}_{\mathrm{glp}}^{\top}
\tanh
\left(
\mathbf{W}_{\mathrm{glp}}\mathbf{h}_{c_n}^{\mathrm{ref}}
+
\mathbf{U}_{\mathrm{glp}}\mathbf{q}^{(0)}
\right),
\quad n=1,\ldots,N_p,
\]
\[
\beta_n^{\mathrm{glp}}
=
\frac{\exp(e_n^{\mathrm{glp}})}
{\sum_{r=1}^{N_p}\exp(e_r^{\mathrm{glp}})},
\quad
\mathbf{r}
=
\sum_{n=1}^{N_p}
\beta_n^{\mathrm{glp}}
\mathbf{h}_{c_n}^{\mathrm{ref}}.
\]
The pointer query is updated using the glimpse summary:
\[
\mathbf{q}^{(1)}
=
\operatorname{MLP}_{r}
\left(
[
\mathbf{q}^{(0)};
\mathbf{r}
]
\right).
\]

Finally, each candidate is assigned a pointer logit:
\[
o_n
=
\mathbf{v}_{o}^{\top}
\tanh
\left(
\mathbf{W}_{o}\mathbf{h}_{c_n}^{\mathrm{ref}}
+
\mathbf{U}_{o}\mathbf{q}^{(1)}
\right),
\quad n=1,\ldots,N_p.
\]
The logits $\{o_n\}_{n=1}^{N_p}$ are normalized over the current candidate set to obtain the receiver distribution defined in Section~\ref{sec:problem_formulation}.

\subsection{Training Objective}
\label{subsec:training_objective}

The model is trained with candidate-level negative log-likelihood:
\[
\mathcal{L}
=
-
\frac{1}{|\mathcal{D}|}
\sum_{p \in \mathcal{D}}
\log
P(y_p \mid p,\mathcal{C}_p),
\]
where $\mathcal{D}$ is the training set, $y_p$ is the ground-truth receiver index, and $\mathcal{C}_p$ is the candidate set for pass $p$. 
We use accuracy as the primary evaluation metric.

The per-pass inference complexity is
$O\!\bigl((NM+N)d + L^2d + NLd + (N+M+L)d^2\bigr)$,
where $N$, $M$, $L$, and $d$ denote the numbers of receiver candidates, visible opponents, historical events, and the hidden dimension, respectively. The $(NM+N)d$ term comes from graph message passing, $L^2d$ from temporal self-attention over historical events, $NLd$ from candidate--history attention, and $(N+M+L)d^2$ from feature projections, fusion layers, and pointer-based scoring. Since the graph is small and $L \leq K$, inference remains manageable. During training, the same forward computation is used with standard backpropagation overhead.

\section{Experiments}

\subsection{Datasets}

We use publicly available StatsBomb Open Data with 360 freeze-frame annotations. 
Each freeze frame provides the visible pitch area and player locations for selected events. 
Our prediction targets are pass events from competitions with available 360 data. Event context and possession history are constructed from the event stream, including preceding events within the same possession. Since StatsBomb 360 is reconstructed from broadcast footage, a freeze frame may not include all 22 players on the pitch. Instead, it only contains players visible within the camera view, together with the corresponding visible area. 
Thus, receiver selection is performed over visible teammate candidates under partial observation. 
Table~\ref{tab:statsbomb_360_pass_summary} summarizes the six datasets used in our experiments.

\begin{table}[th]
\vspace{-0.85cm}
\centering
\caption{Summary of public StatsBomb datasets with available 360 data}
\label{tab:statsbomb_360_pass_summary}
\resizebox{\textwidth}{!}{
\begin{tabular}{llccccc}
\toprule
\textbf{Competition} & 
\textbf{Season} &
\textbf{Abbr.} & 
\textbf{Match} & 
\textbf{Pass} & 
\textbf{\makecell{Avg.\\Pass}} & 
\textbf{\makecell{Visible\\Pass}} \\
\midrule
1. Bundesliga        & 2023/24 & BL 23/24       & 34 & 39214 & 1153.4 & 25472 \\
FIFA World Cup       & 2022      & FIFA WC 2022   & 64 & 68515 & 1070.5 & 40952 \\
UEFA Euro            & 2020      & Euro 2020      & 51 & 54819 & 1074.9 & 32677 \\
UEFA Euro            & 2024      & Euro 2024      & 51 & 53890 & 1056.7 & 34373 \\
UEFA Women's Euro    & 2025      & WEuro 2025     & 31 & 29108 & 939.0  & 16005 \\
Women's World Cup    & 2023      & FIFA WWC 2023  & 64 & 59837 & 935.0  & 32526 \\
\bottomrule
\end{tabular}
}
\vspace{-0.85cm}
\end{table}

\subsection{Experimental Setup}

For each dataset, we first select 15\% of matches as the test set and hold out all pass samples from these matches for final evaluation. 
Samples from the remaining matches are split at the sample level into training and validation sets with an 8:2 ratio. 
The validation set is used for model selection and early stopping. 
All methods are evaluated by accuracy, which measures whether the predicted receiver matches the ground-truth receiver among the visible candidates.

All models are trained with Adam using a learning rate of $10^{-3}$, a batch size of 32, a hidden dimension of 64, 4 attention heads, and dropout of 0.1. 
Early stopping is based on validation accuracy with a patience of 10 epochs, and the best validation checkpoint is used for final test evaluation.

\subsection{Experimental Results}

\subsubsection{Comparison with Baselines}

HPGPN is compared with representative baselines under the same input constraints as un-xPass: Closest Teammate, SoccerMap, and XGBoost. 
Closest Teammate is a heuristic baseline, SoccerMap is a pitch-based spatial deep learning baseline, and XGBoost is a strong feature-based baseline. Table~\ref{tab:main_results} reports pass receiver selection accuracy across six datasets.

HPGPN achieves the best accuracy on all datasets, consistently outperforming heuristic, spatial deep learning, and feature-based baselines. 
Among the baselines, XGBoost is generally the strongest competitor, indicating that handcrafted spatial and contextual features remain useful for receiver selection. 
Compared with XGBoost, HPGPN improves the average accuracy from 0.5387 to 0.5718, yielding an absolute gain of 3.31 percentage points. 
The largest gains are observed on Euro 2020 and WEuro 2025, with improvements of 5.55 and 4.42 percentage points, respectively. 
In contrast, the smaller gain on BL 23/24 suggests that the benefit of possession-aware graph pointer modeling may depend on dataset characteristics, such as competition style, visible-player distribution, and candidate-set structure. Overall, these results show that receiver selection cannot be fully captured by nearest-teammate heuristics or pitch-based spatial representations alone. 
By jointly modeling player interactions, event-level context, and possession history, HPGPN captures additional decision cues and better selects the intended receiver from a variable set of visible candidates.

\begin{table}[htbp]
\vspace{-0.9cm}
\centering
\caption{Main results of pass receiver selection accuracy}
\label{tab:main_results}
\setlength{\tabcolsep}{4pt}
\renewcommand{\arraystretch}{1.12}
\resizebox{0.92\linewidth}{!}{
\begin{tabular}{@{}lccccccc@{}}
\toprule
\textbf{Method} 
& \makecell{\textbf{BL}\\\textbf{23/24}}
& \makecell{\textbf{FIFA WC}\\\textbf{2022}}
& \makecell{\textbf{Euro}\\\textbf{2020}}
& \makecell{\textbf{Euro}\\\textbf{2024}}
& \makecell{\textbf{WEuro}\\\textbf{2025}}
& \makecell{\textbf{FIFA WWC}\\\textbf{2023}}
& \textbf{Avg.} \\
\midrule
Closest teammate 
& 0.3283 
& 0.4055 
& 0.3676 
& 0.3707 
& 0.3955 
& 0.4024
& 0.3783 \\

Soccermap 
& 0.3706 
& 0.4281 
& 0.4224 
& 0.4307 
& 0.3740 
& 0.4204
& 0.4077 \\

XGBoost 
& 0.5118 
& 0.5786 
& 0.5059 
& 0.5669 
& 0.5233 
& 0.5457
& 0.5387 \\
\midrule
HPGPN
& \textbf{0.5301} 
& \textbf{0.6007}
& \textbf{0.5614}
& \textbf{0.6001}
& \textbf{0.5675} 
& \textbf{0.5710}
& \textbf{0.5718} \\
\bottomrule
\end{tabular}
}
\vspace{-1.2cm}
\end{table}

\subsubsection{Cross-Dataset Transferability}

\begin{wrapfigure}{r}{0.50\textwidth}
    \centering
    \vspace{-15pt}
    \includegraphics[width=0.5\textwidth]{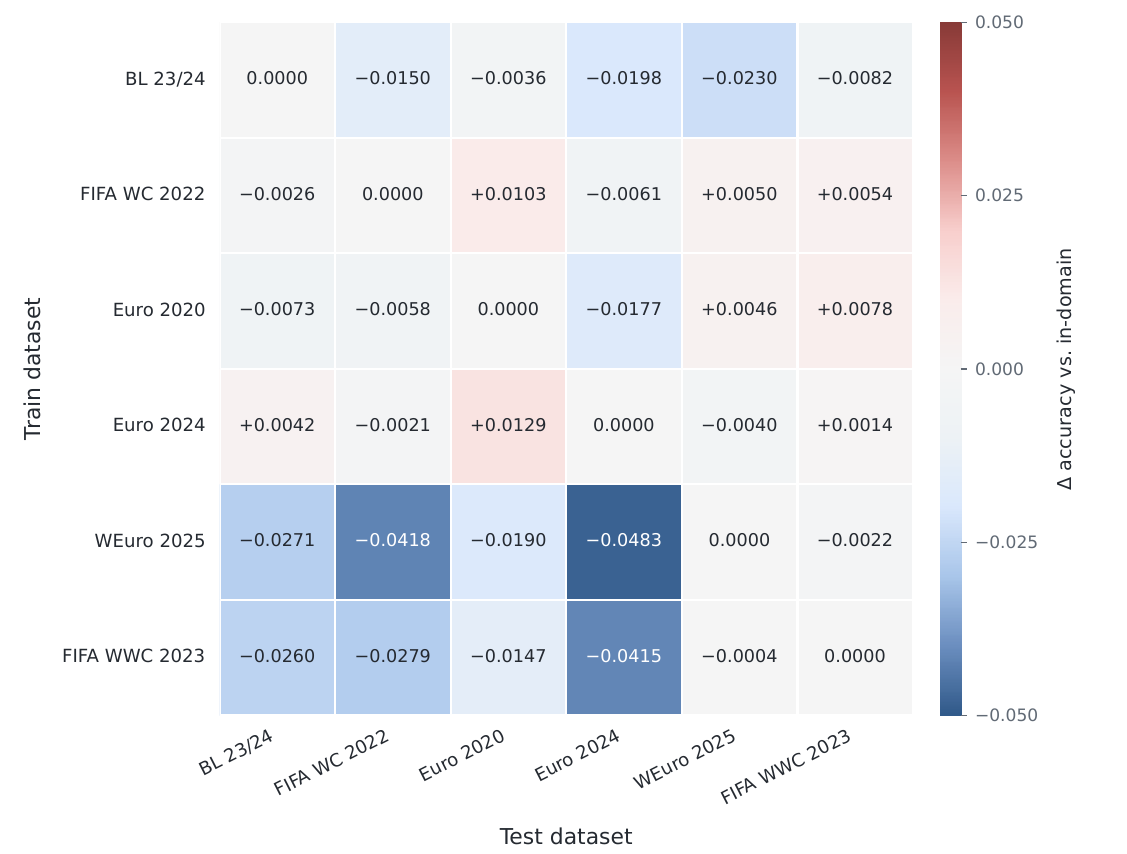}
    \caption{Cross-dataset transfer accuracy difference relative to in-domain training.}
    \label{fig:cross_dataset_transfer_delta}
    \vspace{-10pt}
\end{wrapfigure}

To evaluate cross-competition transferability, we train HPGPN on one source dataset and directly test it on other target datasets without target-domain fine-tuning. 
Fig.~\ref{fig:cross_dataset_transfer_delta} reports the accuracy difference relative to the target in-domain baseline, defined as $\Delta_{s \rightarrow t}=\mathrm{Acc}(s \rightarrow t)-\mathrm{Acc}(t \rightarrow t)$.
Near-zero off-diagonal values indicate preserved transfer performance, while positive values indicate that transfer outperforms the target in-domain model. Overall, the off-diagonal entries do not collapse, showing that HPGPN generalizes to unseen competitions. 
Several settings are close to or even better than the target in-domain baseline. 
Notably, effective transfer is observed within men's competitions and from men's to women's competitions, suggesting that HPGPN captures receiver-selection patterns beyond dataset-specific correlations. 
In contrast, transfers from women's to men's competitions show larger drops, indicating an asymmetric transfer pattern across competition groups.

\subsubsection{Sensitivity to Maximum History Window Length}

\begin{figure}[htbp]
\vspace{-0.2cm}
    \centering

    \begin{subfigure}{0.475\textwidth}
        \centering
        \includegraphics[width=\linewidth]{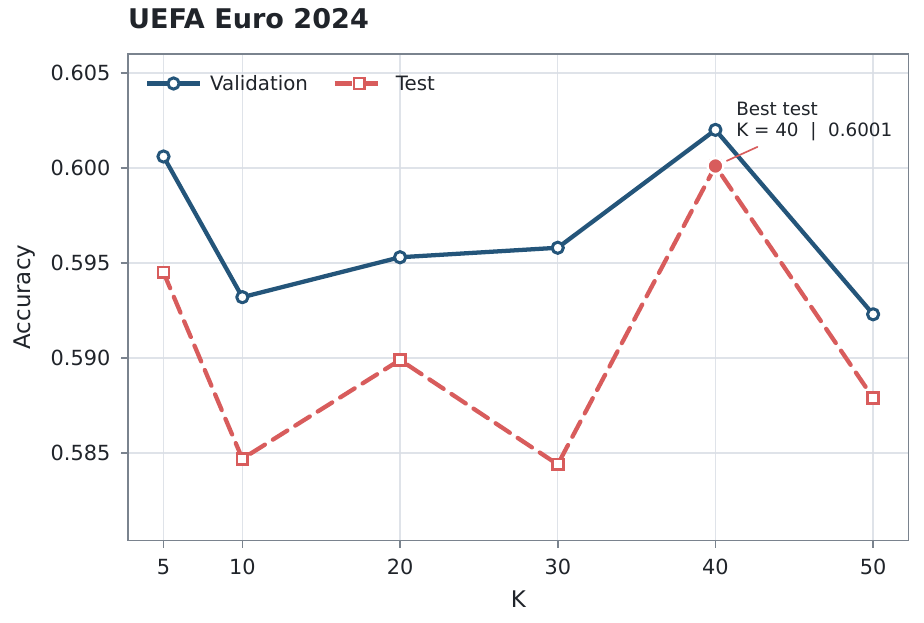}
        %\caption{UEFA Euro 2024}
        \label{fig:history_length_euro2024}
    \end{subfigure}
    \hfill
    \begin{subfigure}{0.475\textwidth}
        \centering
        \includegraphics[width=\linewidth]{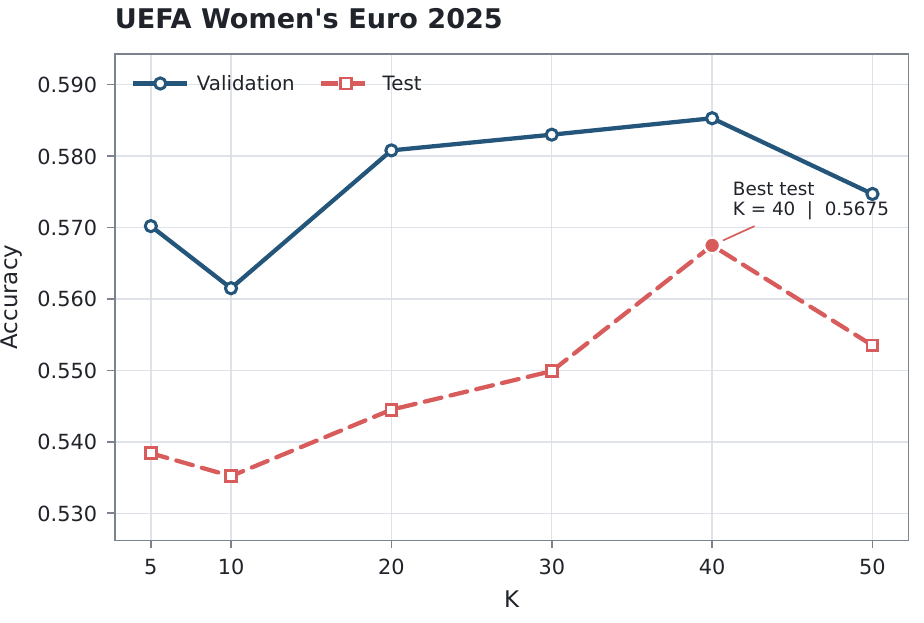}
        %\caption{UEFA Women's Euro 2025}
        \label{fig:history_length_womens_euro2025}
    \end{subfigure}

    \caption{Sensitivity of HPGPN to the maximum history window length $K$ on UEFA Euro 2024 (left) and UEFA Women's Euro 2025 (right).}
    \label{fig:history_length_sensitivity}
\vspace{-0.9cm}
\end{figure}

We evaluate HPGPN with different maximum history window lengths $K$, which set an upper bound on the number of preceding possession events retained. 
Fig.~\ref{fig:history_length_sensitivity} reports validation and test accuracy on Euro 2024 and WEuro 2025. The best test accuracy is achieved at $K=40$ on both datasets, reaching 0.6001 and 0.5675, respectively. Smaller windows provide insufficient possession context, whereas larger windows may introduce less relevant earlier events. These results suggest that possession history is beneficial, but the temporal window should balance contextual information and historical noise.
We therefore set $K=40$ in the experiments.

\subsubsection{Performance under Different Candidate Set Sizes}

\begin{figure}[t]
%\vspace{-0.8cm}
    \centering
    \includegraphics[width=0.9\linewidth]{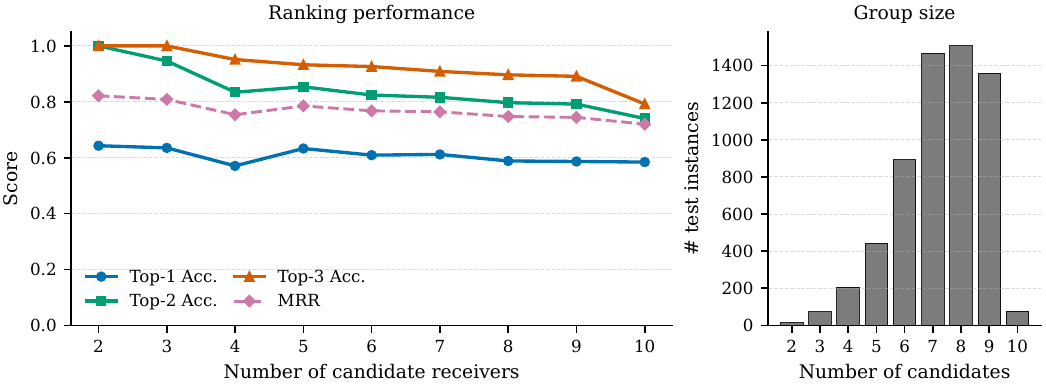}
    \caption{Candidate-set size analysis on the UEFA Euro 2024 test set: ranking performance (left) and sample distribution (right).}
    \label{fig:candidate_count_analysis}
    \vspace{-0.2cm}
\end{figure}

Using Euro 2024 as an example, Fig.~\ref{fig:candidate_count_analysis} reports ranking performance and sample distribution by candidate-set size. 
Performance drops as candidate sets grow, reflecting increased ambiguity among plausible receivers. 
However, Top-2/Top-3 accuracy and MRR remain strong across common sizes, showing that HPGPN often ranks the true receiver near the top even when the Top-1 prediction is incorrect. 
Since most samples contain six to nine visible candidates, medium-sized candidate sets dominate the overall trend. 
These results demonstrate that HPGPN handles variable-size candidate sets and provides informative rankings beyond Top-1 prediction.

\subsection{Ablation Study}

We conduct ablation studies on UEFA Euro 2024 and UEFA Women's Euro 2025 to analyze the contribution of major components in HPGPN. 
All variants use the same experimental setup, and we report receiver-selection accuracy.

\subsubsection{Fixed Event Context and Dynamic Possession History}
Table~\ref{tab:ablation_context_history} analyzes the contextual components of HPGPN. 
Removing the fixed event context consistently reduces accuracy, confirming the value of local event-level cues.  
Removing only the fixed short-term history $a_{t-2}$ and $a_{t-1}$ also reduces accuracy, suggesting that compact recent-action context complements the dynamic possession history.

Removing the dynamic possession-history encoder degrades performance, showing that possession-level temporal information provides useful cues beyond the current pass graph and fixed context. 
Within this encoder, both all-event and pass-only history contexts are beneficial, capturing possession development and previous passing actions, respectively. Performance also drops when history metadata or contextualized possession-history tokens are removed, indicating the importance of availability, length, and temporal contextualization for variable-length histories. Finally, modeling pass history without all-event contextualization or using a disjoint non-pass/pass design performs worse than the full model. 
This suggests that historical passes should be interpreted within broader possession evolution and that pass events should remain part of the all-event branch.

\begin{table}[htbp]
\vspace{-0.7cm}
\centering
\caption{Ablation study of event context and possession history components.}
\label{tab:ablation_context_history}
\setlength{\tabcolsep}{8pt}
\renewcommand{\arraystretch}{1.15}
\resizebox{\textwidth}{!}{
\begin{tabular}{lcc}
\toprule
\textbf{Variants} 
& \makecell{\textbf{UEFA Euro 2024}}
& \makecell{\textbf{UEFA WEuro 2025}} \\
\midrule
w/o fixed event context
& 0.5607 
& 0.5345 \\

w/o overlapping $a_{t-2}$/$a_{t-1}$ event context
& 0.5652 
& 0.5416  \\[5pt]

w/o dynamic possession history  
& 0.5864 
& 0.5406 \\

w/o all-event history context
& 0.5962 
& 0.5488  \\

w/o pass-only history context
& 0.5907 
& 0.5284  \\

w/o history metadata 
& 0.5819 
& 0.5474  \\

w/o contextualized possession-history tokens
& 0.5915 
& 0.5427  \\

w/o surrounding event context for pass history
& 0.5882 
& 0.5377  \\

w/o overlapping all-event/pass branches 
& 0.5733 
& 0.5492  \\
\midrule
HPGPN (ours)
& \textbf{0.6001}
& \textbf{0.5675} \\
\bottomrule
\end{tabular}
}
\vspace{-0.9cm}
\end{table}

\subsubsection{GNN Backbone}
Table~\ref{tab:gnn_backbone} compares different GNN backbones for encoding the current pass graph. GraphSAGE achieves the best performance on both datasets, outperforming GCN and GAT. This suggests that its neighborhood aggregation scheme is better suited for modeling passer--candidate--defender interactions. We therefore use GraphSAGE as the graph encoder in HPGPN.

\subsubsection{Receiver Selection Head}

Table~\ref{tab:pointer_head} compares receiver selection heads. 
The MLP scorer evaluates candidates independently, while the pointer head scores candidates with a context-dependent query. 
The glimpse pointer head summarizes the candidate set before scoring and achieves the best performance on both datasets. 
This suggests that receiver selection benefits from candidate comparison under the same pass context, rather than independent candidate scoring.

\begin{table}[t]
\centering
\scriptsize
\renewcommand{\arraystretch}{1.10}

\begin{minipage}[t]{0.44\textwidth}
\centering
\captionof{table}{GNN backbone}
\label{tab:gnn_backbone}
\setlength{\tabcolsep}{3pt}
\begin{tabular*}{\linewidth}{@{\extracolsep{\fill}}lcc@{}}
\toprule
\textbf{Method} 
& \makecell{\textbf{Euro}\\\textbf{2024}}
& \makecell{\textbf{WEuro}\\\textbf{2025}} \\
\midrule
GCN 
& 0.5897 
& 0.5366 \\
GAT 
& 0.5902 
& 0.5402 \\
\midrule
GraphSAGE (ours)
& \textbf{0.6001}
& \textbf{0.5675} \\
\bottomrule
\end{tabular*}
\end{minipage}
\hfill
\begin{minipage}[t]{0.52\textwidth}
\centering
\captionof{table}{Receiver selection head}
\label{tab:pointer_head}
\setlength{\tabcolsep}{3pt}
\begin{tabular*}{\linewidth}{@{\extracolsep{\fill}}lcc@{}}
\toprule
\textbf{Method} 
& \makecell{\textbf{Euro}\\\textbf{2024}}
& \makecell{\textbf{WEuro}\\\textbf{2025}} \\
\midrule
MLP scorer 
& 0.5986 
& 0.5459 \\
Pointer head 
& 0.5957
& 0.5503 \\
\midrule
Glimpse pointer head (ours)
& \textbf{0.6001}
& \textbf{0.5675} \\
\bottomrule
\end{tabular*}
\end{minipage}
\vspace{-0.3cm}
\end{table}

\section{Conclusion}

In this paper, we proposed HPGPN, a hierarchical possession-aware graph pointer network for pass receiver selection under the partial-observation setting of StatsBomb data. 
HPGPN jointly models the current pass graph, fixed event context, and dynamic possession history, and uses a glimpse pointer head to select the intended receiver from a variable set of visible candidate teammates. 
Experiments on multiple public datasets show that HPGPN consistently outperforms heuristic, spatial deep learning, and feature-based baselines, while ablation studies validate the contribution of its key components. These results show that receiver selection benefits from structured modeling of player interactions, event-level context, and possession-level temporal information. 
Future work will extend the framework to basketball and other team sports, where decision-making also depends on dynamic multi-agent interactions and evolving possession context.

\subsubsection{Acknowledgements.}
 This work was supported by the National Natural Science Foundation of China (Grant No.~62506018) and was also partially supported by the Open Fund of Beijing Key Laboratory of Interdisciplinary Intelligent Technologies in Sports Medicine and Engineering. We thank Changhong Jin (University College Dublin) and the anonymous reviewers for their valuable comments and constructive suggestions.

\bibliographystyle{splncs04}
\bibliography{mybibliography}

@book{fujii2025machine,
  title={Machine learning in sports: open approach for next play analytics},
  author={Fujii, Keisuke},
  year={2025},
  publisher={Springer Nature}
}

@inproceedings{dauxais2018predicting,
  title={Predicting pass receiver in football using distance based features},
  author={Dauxais, Yann and Gautrais, Cl{\'e}ment},
  booktitle={International Workshop on Machine Learning and Data Mining for Sports Analytics},
  pages={145--151},
  year={2018},
  organization={Springer}
}

@inproceedings{fournier2018football,
  title={Football pass prediction using player locations},
  author={Fournier-Viger, Philippe and Liu, Tianbiao and Chun-Wei Lin, Jerry},
  booktitle={International Workshop on Machine Learning and Data Mining for Sports Analytics},
  pages={152--158},
  year={2018},
  organization={Springer}
}

@inproceedings{hubavcek2018deep,
  title={Deep learning from spatial relations for soccer pass prediction},
  author={Hub{\'a}{\v{c}}ek, Ond{\v{r}}ej and {\v{S}}ourek, Gustav and {\v{Z}}elezn{\`y}, Filip},
  booktitle={International workshop on machine learning and data mining for sports analytics},
  pages={159--166},
  year={2018},
  organization={Springer}
}

@inproceedings{li2018predicting,
  title={Predicting the receivers of football passes},
  author={Li, Heng and Zhang, Zhiying},
  booktitle={International Workshop on Machine Learning and Data Mining for Sports Analytics},
  pages={167--177},
  year={2018},
  organization={Springer}
}

@inproceedings{vercruyssen2016qualitative,
  title={Qualitative spatial reasoning for soccer pass prediction},
  author={Vercruyssen, Vincent and De Raedt, Luc and Davis, Jesse},
  booktitle={Machine Learning and Data Mining for Sports Analytics (MLSA 2016)@ ECML/PKDD 2016, Riva del Garda, Italy, September 19, 2016},
  year={2016},
  organization={Technical University of Aachen}
}

@article{sanyal2021will,
  title={Who will receive the ball? Predicting pass recipient in soccer videos},
  author={Sanyal, Samriddha},
  journal={Journal of Visual Communication and Image Representation},
  volume={78},
  pages={103190},
  year={2021},
  publisher={Elsevier}
}

@INPROCEEDINGS{honda2022pass,
  author={Honda, Yutaro and Kawakami, Rei and Yoshihashi, Ryota and Kato, Kenta and Naemura, Takeshi},
  booktitle={2022 IEEE/CVF Conference on Computer Vision and Pattern Recognition Workshops (CVPRW)}, 
  title={Pass Receiver Prediction in Soccer using Video and Players’ Trajectories}, 
  year={2022},
  volume={},
  number={},
  pages={3502-3511},
  doi={10.1109/CVPRW56347.2022.00394}}

@article{paneru2024enhancing,
  title={Enhancing soccer pass receiver prediction in broadcast images through advanced deep learning techniques: A comprehensive study on model optimization and performance evaluation},
  author={Paneru, Biplov and Paneru, Bishwash and Poudyal, Ramhari and Poudyal, Khem},
  journal={Journal of Soft Computing Exploration},
  volume={5},
  number={2},
  pages={115--121},
  year={2024}
}

@INPROCEEDINGS{kaneko2024augmenting,
  author={Kaneko, Takeshi and Kawakami, Rei and Naemura, Takeshi and Inoue, Nakamasa},
  booktitle={2024 IEEE/CVF Conference on Computer Vision and Pattern Recognition Workshops (CVPRW)}, 
  title={Augmenting Pass Prediction via Imitation Learning in Soccer Simulations}, 
  year={2024},
  volume={},
  number={},
  pages={3194-3203},
  doi={10.1109/CVPRW63382.2024.00325}}

@article{rahimian2026temporal,
  title={Temporal Graph Network Framework for Quantifying Pass Reception Probabilities Against Defensive Structures},
  author={Rahimian, Pegah and Davis, Jesse and Toka, Laszlo},
  journal={Machine Learning},
  volume={115},
  number={1},
  pages={6},
  year={2026},
  publisher={Springer}
}

@inproceedings{fernandez2020soccermap,
  title={Soccermap: A deep learning architecture for visually-interpretable analysis in soccer},
  author={Fern{\'a}ndez, Javier and Bornn, Luke},
  booktitle={Joint European Conference on Machine Learning and Knowledge Discovery in Databases},
  pages={491--506},
  year={2020},
  organization={Springer}
}

@inproceedings{robberechts2023xpass,
  title={un-xpass: Measuring soccer player's creativity},
  author={Robberechts, Pieter and Van Roy, Maaike and Davis, Jesse},
  booktitle={Proceedings of the 29th ACM SIGKDD conference on knowledge discovery and data mining},
  pages={4768--4777},
  year={2023}
}

@article{vinyals2015pointer,
  title={Pointer networks},
  author={Vinyals, Oriol and Fortunato, Meire and Jaitly, Navdeep},
  journal={Advances in neural information processing systems},
  volume={28},
  year={2015}
}

@article{ma2019combinatorial,
  title={Combinatorial optimization by graph pointer networks and hierarchical reinforcement learning},
  author={Ma, Qiang and Ge, Suwen and He, Danyang and Thaker, Darshan and Drori, Iddo},
  journal={arXiv preprint \href{https://arxiv.org/abs/1911.04936}{arXiv:1911.04936}},
  year={2019}
}

@inproceedings{
kipf2017semi,
title={Semi-Supervised Classification with Graph Convolutional Networks},
author={Thomas N. Kipf and Max Welling},
booktitle={International Conference on Learning Representations},
year={2017},
url={https://openreview.net/forum?id=SJU4ayYgl}
}

@article{hamilton2017inductive,
  title={Inductive representation learning on large graphs},
  author={Hamilton, Will and Ying, Zhitao and Leskovec, Jure},
  journal={Advances in neural information processing systems},
  volume={30},
  year={2017}
}

@inproceedings{
velivckovic2018graph,
title={Graph Attention Networks},
author={Petar Veličković and Guillem Cucurull and Arantxa Casanova and Adriana Romero and Pietro Liò and Yoshua Bengio},
booktitle={International Conference on Learning Representations},
year={2018},
url={https://openreview.net/forum?id=rJXMpikCZ},
}
\end{document}